\documentclass[lettersize,journal]{IEEEtran}
\usepackage{amsmath,amsfonts,amssymb}
\usepackage{algorithm}
\usepackage{algpseudocode}
\usepackage{algorithmicx}
\usepackage{array}
\usepackage[caption=false,font=normalsize,labelfont=sf,textfont=sf]{subfig}
\usepackage{textcomp}
\usepackage{stfloats}
\usepackage{bbm}
\usepackage{url}
\usepackage{multirow}
\usepackage{verbatim}
\usepackage{graphicx}
\usepackage{cite}
\usepackage{enumitem}
\begin{document}
	\bstctlcite{IEEEexample:BSTcontrol}
	
	\title{Tool-to-Tool Matching Analysis Based Difference Score Computation Methods for Semiconductor Manufacturing}
	
	\author{Sameera Bharadwaja H., Siddhrath Jandial, Shashank S. Agashe, Rajesh Kumar Reddy Moore, Youngkwan Kim
		\thanks{Sameera Bharadwaja H., Siddhrath Jandial, Shashank Srikanth Agashe, and Rajesh Kumar Reddy are with Smart Equipment Solution Group (SESG), Samsung Semiconductor India Research (SSIR), Bangalore, India (email: s.hayavadana@samsung.com; siddharth.j@samsung.com; shashank.a@samsung.com; rajesh.moore@samsung.com).}
		\thanks{Youngkwan Kim is with Mechatronics Research (MR), Device Solutions Research (DSR), Samsung Electronics Co., South Korea (email: yk313.kim@samsung.com).}}

\maketitle

\begin{abstract}
	We consider the problem of \emph{tool-to-tool matching (TTTM)}, also called, \emph{chamber matching} in the context of a semiconductor manufacturing equipment. Traditional TTTM approaches utilize static configuration data or depend on a golden reference which are difficult to obtain in a commercial manufacturing line. Further, existing methods do not extend very well to a \emph{heterogeneous setting}, where equipment are of different make-and-model, sourced from different equipment vendors. We propose novel \emph{TTTM analysis pipelines} to overcome these issues. We hypothesize that a mismatched equipment would have higher variance and/or higher number of modes in the data. Our best univariate method achieves a correlation coefficient $>0.95$ and $>0.5$ with the variance and number of modes, respectively showing that the proposed methods are effective. Also, the best multivariate method achieves a correlation coefficient $>0.75$ with the top-performing univariate methods, showing its effectiveness. Finally, we analyze the sensitivity of the multivariate algorithms to the algorithm hyper-parameters.
\end{abstract}


\section{Introduction}
\label{sec:Introduction}
\IEEEPARstart{I}{ncrease} in market demands for high quality semiconductor chips has compelled semiconductor industries to adopt measures to increase utilization and yield of the production line in the semiconductor fabrication plants (Fab)~\cite{Burkacky_Dragon_Lehmann_2022}. In a large scale Fab, it is important to ensure that the production line yield is consistently high. The Fab line is comprised of various equipment catering to one or more of 500-1200 steps that are involved in the chip manufacturing. Therefore, the problem of production line consistency can be decomposed down into a similar problem, but at the \emph{tool} level i.e., per equipment or equipment-chamber, in case of a cluster-equipment. Further, it is often important to characterize the consistency of an equipment at the sub-system level for advanced process control (APC)~\cite{Sarfaty_Paik_Parikh_2002}. Consistency quantification and monitoring is achieved using tool-to-tool matching (TTTM), also called, chamber-matching. Static tool configuration matching fail to take into account the effects of the part aging, dynamic recipes, product mix, to name a few and hence, perform sub-optimally in practice. Various literature works propose data-driven approach to overcome the disadvantage mentioned above.

\subsection{Prior Work}
\label{sec:prior work}
The authors in~\cite{Chiou_Tso_2003, Vaid_Solecky_Su_2015} propose a method in which the metrology data distribution is analyzed and a \emph{degree of matching} is quantified for each tool. The authors in~\cite{Pan_Wong_Jang_2012} propose statistical hypothesis testing approaches like F-test, IQR-test and linear regression to assess the statistical variables of the wafer quality data. A detailed overview of the statistical chart and hypothesis test-based techniques are summarized in~\cite{Feyz_2016}. However, metrology operations are time-consuming and often, destructive to the wafers~\cite{Orji_Badaroglu_2018}. Therefore, metrology data is collected only at critical steps, for sampled wafers.

An alternative, inspired by the success of virtual metrology (VM) pipelines over past few years, is that one can rely on matching the process data~\cite{Kang_Lee_2009, Chang_Kang_2006, Dreyfus_May_2022}. We briefly describe the concept of \emph{process data matching}. In any modern equipment, multiple in-situ sensors collect large quantity of process data in the form of multivariate time-series~\cite{Han_Min_Ma_2023}, where each variate represents the data from one sensor. The sensors measure the status of physical variables in the chamber, like, \emph{Chiller Flow}, \emph{Gas Flow} (Helium, Argon, etc.), \emph{Clamp Pressure}, \emph{Temperature}, \emph{Valve position of the mass flow controller (MFC)}, \emph{RF bias}, \emph{Impedance}, to name a few. The collected time-series data represents the tool health and has near real-time validity to the tool’s actual state. Often, particular set of sensors form necessary and sufficient trackers for the tool's sub-system. For instance, a set of temperature sensors can accurately represent the health/ state of the processing chamber's temperature control sub-system~\cite{Kevin_Turner_Jackson_2000}.

The author in~\cite{Banna_2020} presents an adaptive chamber matching pipeline which relies on VM and on-board metrology (OBM) component. In the above methods, identification of a \emph{golden reference} chamber or wafer is crucial. The authors in~\cite{Marino_Rossi_2017} propose an unsupervised chamber matching method based on matching the shape of the process-chamber sensor measured time-series data over the course of the manufacturing step (E.g., Etching, Deposition, to name a few). However, the degree of variation in the shape of the sensor data alone does not directly relate to how the quality of the process, and hence the wafer behaves due to the presence of feedback control loops in the manufacturing equipment. Further, the physical parameters measured by various sensors like pressure gauges, temperature sensors, residual gas analyzers, gas flow monitors, plasma health correlate to varying degrees as stipulated by the operating conditions. Therefore, multivariate analysis (MVA) methods for process data matching by taking into account the complex relationship between the variates are necessary.

The authors in~\cite{Kye_Han_Choi_2013} propose variable selection techniques like singular value analysis and relative gain array methods for plasma etch chamber matching. The authors in~\cite{Chen_Chang_2013} disclose a pipeline comprising of various MVA methods like principle component analysis (PCA), partial least-squares discriminant analysis model (PLS-DA). PCA treats the variance of the data variates as the \emph{information} and relies of decomposing the variates into various independent basis directions in a vector space. On the other hand, PLA-DA could be treated as a supervised extension of PCA.

Recently, neural network architectures which can analyze and account for non-linear information and correlation content in the multivariate data have been applied by the authors in~\cite{Heng_Liao_2021}. The neural network approach presented in~\cite{Heng_Liao_2021} relies on learning a latent representation of the multivariate time-series data by using an auto-encoder-like structure and minimizes the reconstruction error between the original and the reconstructed data. The data from a chamber that is within specifications is used to train the model. Thereafter, the error score is computed by feeding the data from a chamber-of-interest and checked if the score falls within a threshold or not. Other notable literature on chamber or tool matching includes the works by the authors in~\cite{James_Manjunath_Jimmy_2014, Aabir_Blue_Yugma_2019} in the direction of run-to-run (R2R) control. A typical neural network model's layers mix the raw sensor data in a non-linear manner. Therefore, the behavior and outcome is not explainable in terms of the original sensor variate, in a straightforward manner. Other notable works on TTTM include tool-specific methods proposed by the authors in~\cite{Zhou_Kim_Young_2016, Ng_Zhou_2003}. However, these methods do not extend to a heterogeneous setting\footnote{A set of tools used in the same processing step can differ in the model and hence, in the physical configurations etc. due to being sourced from different manufacturer or suppliers, software and part upgrade etc. Similarly, aging effect of the tools can vary due to different mix of recipes under which the tools operate over a period of time in a modern commercial Fab.}; typical characteristics any modern Fab would exhibit. In addition, one of the major challenges in using the raw process sensor data is the big-data problem, with more than few hundred process sensors, typically present in the chamber. The number of chamber-sensors will only increase as the chip size further reduces demanding for more accurate fault detection and classification (FDC) and hence, more reliable advanced process control (APC)~\cite{Sarfaty_Paik_Parikh_2002, Kevin_Turner_Jackson_2000}.

\subsection{Contributions}
\label{sec:contribution_takeawway}
In light of the gaps identified above, we propose two sets of computationally efficient and universal, i.e., tool-independent, pipelines, whose output is a \emph{TTTM difference score} assigned to each tool under consideration. The first set caters to univariate analysis, where the difference scores are obtained at sensor level. The second set comprises of graph neural network (GNN)-based multivariate TTTM analysis algorithms. As the name suggests, the difference scores are proportional to the degree of inconsistency i.e., deviation from the characteristics of other tools, on an average. As we shall see in Sec.~\ref{sec:Motivation for Multivariate Analysis}, our multivariate pipelines extend to heterogeneous settings thereby overcoming the drawbacks of the above methods.

In order to overcome the issue of huge data volume of the raw sensor time-series (trace), we propose a dimension-reducing encoding operation in the pre-processing step (see Sec.~\ref{sec:Trend_Removal} for more details). The pre-processed data is called the trace-summary (T-SUM) data. The advantage of using the derived T-SUM data is 1) the reduced data volume resulting in efficient processing; and 2) implicit modeling of the long-term aging effects, which form one of the most informative features for TTTM are more evident.

The main contributions of our work are as follows:
\begin{enumerate}
	\item We propose three algorithms for computing the TTTM difference score in univariate setting, where each sensor is independently analyzed. That is, the score is first computed at the sensor level and an aggregation operation is then applied to obtain the final tool-level score. The intermediate sensor-scores can serve as a crucial input to the APC control sub-routines.
	\item We propose two GNN-based multivariate TTTM analysis pipelines. Specifically, we learn the \emph{graph model} of each tool from the corresponding multivariate T-SUM data using graph attention networks (GAT)~\cite{Nurul_Yeahia_Ripon_2021, Zhou_Cui_2020}. The self-attention weights in GATs ensure that the learnt model is robust to noise in the data and unreliable node features~\cite{Petar_Yoshua_2017}. Finally, we formulate a graph-edit distance to capture the deviation in the graph models between any pair of tools. Pair-wise tool-scores are aggregated to obtain the TTTM difference score for each tool.
	\item We propose and illustrate a standard operating procedure (SOP) for monitoring the trend of the TTTM difference score at both sensor and tool level, with an application to track the deviations in the Fab, in the context of predictive maintenance (PdM) scheduling~\cite{Iskandar_Moyne_2015}. The SOP is described by a case study: We perform the TTTM difference score analysis on the datasets collected across consecutive months using our proposed univariate scoring pipeline. The observations related to the SOP is described and contrasted, in the context of tool/ tool-part (sensor) behavior monitoring and PdM scheduling. Further, ranking by the difference score gives the perception of which sensors have higher probability to be anomalous. Therefore, our scoring model can be used for early-stage anomaly detection and alarm tracking.
\end{enumerate}

The main takeaway of our work is that computing and monitoring TTTM difference scores at part (sensor) and/ or tool-level is a necessary and sufficient metric for monitoring the process consistency. Further, we illustrate the SOP for PdM scheduling, early-stage anomaly detection and alarm tracking. We also show that the GNN architectures are an efficient way of modeling the sensor interaction and perform multivariate TTTM analysis in a data driven manner. Further, we use the difference score and track the delta change in the score over time. Therefore, GNN-based multi-variate TTTM extends the analysis to heterogeneous setting. As per the authors' knowledge, the GNN architecture has not been explored for multivariate TTTM analysis. We bridge this gap in this paper and also, propose a graph edit distance to model the TTTM difference score in a multivariate setting.

\emph{Notations:} Let $\mathbb{R}$ and $\mathbb{Z}$ denote the set of real numbers and integers, respectively. Let $\mathcal{N}(\mu, \sigma^2)$ and $\text{Unif}(a,b)$ denote the Gaussian distribution with mean, $\mu$, and variance, $\sigma^2$; and a uniform distribution in $[a, b]$, respectively. $\mathbb{E}[f(X)]$ denotes the expectation of $f(X)$ over the distribution, $f_X$, of the random variable, $X$. We use lowercase boldface, uppercase boldface and uppercase calligraphic letters to denote vectors, matrices and sets, respectively. The cardinality of a set $\mathcal{X}$ is denoted by $|\mathcal{X}|$. We denote the $(i.j)$th element of a matrix $\mathbf{A}$ by $A(i,j)$. On similar lines, the $i$th row and $j$th column of matrix $\mathbf{A}$ is denoted by $\mathbf{A}(i,:)$ and $\mathbf{A}(:,j)$ respectively. Finally, $\|\mathbf{x}\|_p,~p > 0$ denotes the $l_p$ norm of a vector, $\mathbf{x}$. 

\section{Dataset Preparation}
\label{sec:Dataset Preparation}
In this section, we describe the dataset that drives the proposed TTTM difference scoring pipeline. We start with a collection of $Q$ tools indexed by $\{1, \ldots, Q\} \equiv [Q]$. Denote the set of sensors in the $q$th tool by $\mathcal{S}_q$, for $q \in [Q]$ such that $|\mathcal{S}_q| = N_q$. In the $\bar{k}$th process run, each sensor collects $t_{\bar{k}}$ samples of the associated process parameter~\cite{Robert_2007}. Consolidating the process sensor data (a.k.a. the trace data) of the $q$th tool across all the sensors, we construct $\mathbf{D}_{q\bar{k}}^\prime \in \mathbb{R}^{N_q \times t_{\bar{k}}}$. Finally, collect the trace data across $\bar{K}$ process-runs in $\mathbf{D}_q^\prime = [\mathbf{D}_{q1}^\prime, \mathbf{D}_{q2}^\prime, \ldots, \mathbf{D}_{q\bar{K}}^\prime] \in \mathbb{R}^{N_q \times p_q^\prime},$ where $p_q^\prime = \sum_{\bar{k}=1}^{\bar{K}} t_{\bar{k}}$.

As mentioned in Sec.~\ref{sec:Introduction}, the raw trace data is pre-processed by a dimension-reducing encoder called T-SUM-Trace encoder (TST encoder), denoted by $f_e^{\text{TST}}: \mathbb{R}^{N_q \times p_q^\prime} \to \mathbb{R}^{N_q \times p_q}$, where we set the value of $p_q \leq p_q^\prime$ with $\mathbf{D}_q = f_e^{\text{TST}}\left( \mathbf{D}_q^\prime \right)$. We note that the TST encoder can be positioned as a stream-extract-transform-load (S-ETL) module, which can reduce large time-series data to smaller T-SUM data in near real-time without much storage overhead.

\begin{figure}[!t]
	\centering
	\includegraphics[width=3.4in]{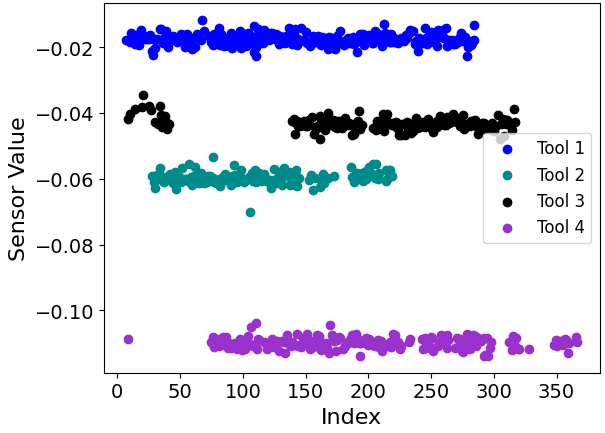}
	\caption{Illustration of the T-SUM Data for one sensor over a duration of $1$ month across $4$ tools.}
	\label{fig:sensor_data}
\end{figure}

In the current work, we use a proprietary deterministic rule-based TST encoder mapping based on statistical features of the trace data. Fig.~\ref{fig:sensor_data} shows a sample T-SUM data for $4$ tools collected over one-month duration.\footnote{The raw trace data is proprietary.} From the figure, we note that the number of samples, i.e., the number of TSUM data collected across different nodes can be different. Further, the duration and the frequency with which the TSUM data is collected is different across the tools. These two aspects make the TTTM difference score computation non-trivial and challenging. In addition, a preventive maintenance (PM) schedule adds to the challenges.

\section{Univariate Scoring Algorithms}
\label{sec:Univariate Scoring Algorithms}
In this Section, we present the proposed univariate scoring algorithms used to compute the difference scores at both the sensor and the tool-level.
\subsection{Clustering Based Scoring}
\label{sec:Clustering_Based_Scoring}
\begin{algorithm}[t!]
	\caption{Clustering-based Scoring Method}
	\label{algo:dbscan_core}
	\begin{algorithmic}[1]
		\Require{$\{\mathbf{D}_q\}_{q=1}^{Q}$, $\{\mathcal{S}_q\}_{q=1}^{Q}$}
		\Ensure{$\mathbf{d}^\text{DBSCAN} \in \mathbb{R}^Q$}
			\State Compute $\mathcal{S} = \cup_{q=1}^{Q} \mathcal{S}_q$ and set $S \gets |\mathcal{S}|$.	
			\State Initialize $p_{\min}$, $\mathbf{d}_{\text{Sensor}}^{\text{DBSCAN}} \in \mathbb{R}^S$.
			\For {$s \in \mathcal{S}$}
				\State Initialize $\mathbf{d}_s$.
				\For {$q \gets 1~\text{to}~Q$}
					\If {$s \in \mathcal{S}_q$}
						\State $i_s \gets \mathcal{S}_q.{\text{IndexOf}}(s)$.
						\State Update $\mathbf{d}_s \gets \text{HorizontalStack}(\mathbf{d}_s, \mathbf{D}_q(i_s, :))$.
					\EndIf
				\EndFor
				\State Compute $\epsilon \gets~{\text{KneeMethod}}\left(\{\mathbf{D}_q\}_{q=1}^{Q}, p_{\min}\right)$~\cite{Rahmah_2016}.
				\State $\text{labels} \gets {\text{Cluster}}(\mathbf{d}_s, p_{\min}, \epsilon)$~\cite{Ester_Kriegel_sander_Xu_1996}.
				\State $\mathcal{C}_{\text{ref}} \gets~\text{IdentifyReferenceCluster}(\mathbf{d}_s, \text{labels})$.
				\State $c_s \gets~\text{ComputeCentroid}(\mathcal{C}_{\text{ref}})$.
				\State $\mathbf{d}_{\text{Sensor}}^{\text{DBSCAN}}(s) \gets \underset{q \in [Q]}{\max}\left\{\sqrt{\frac{\sum_{i=1}^{p_q}|\mathbf{D}_s(i) - c_s|^2}{p_q}}\right\}$.
			\EndFor
			\For {$q \gets 1~\text{to}~Q$}
				\State Set $\mathbf{d}^{\text{DBSCAN}}(q) \gets \frac{\sum_{s \in \mathcal{S}_q}\mathbf{d}_{\text{Sensor}}^{\text{DBSCAN}}(s)}{N_q}$.
			\EndFor
	\end{algorithmic}
\end{algorithm}
The proposed computation algorithm is summarized in Algorithm~\ref{algo:dbscan_core}. We start by collecting all the unique sensors across all the $Q$ tools in a set, $\mathcal{S}$, letting $|\mathcal{S}| = S$. For each sensor, $s \in \mathcal{S}$, we collect the data from all the tools that contain the data for sensor $s$ in a vector, $\mathbf{d}_s$, as shown in steps $5-10$ of Algorithm~\ref{algo:dbscan_core}. A clustering algorithm is then used to identify the clusters in the data as shown in the step $12$.

We use the density-based spatial clustering of applications with noise (DBSCAN) method for clustering the data~\cite{Ester_Kriegel_sander_Xu_1996, Aggarwal_Reddy_2013}. The density-based clustering algorithms identify unique groups/ clusters in the data, based on the idea that a cluster in vector space is a \emph{continuous region of high density}, separated from other such clusters by low density region. Also, the outliers are not erroneously incorporated in a cluster; rather, they are treated as an outlier thereby ensuring the cluster~\emph{purity}. The DBSCAN algorithm requires $p_{\min} > 0$, i.e., the minimum number of data points required to form a legitimate cluster and an $\epsilon > 0$ denoting the maximum distance between any two data points below which they belong to the same neighborhood~\cite{Aggarwal_Reddy_2013}.

In our work, we compute the optimum value of $\epsilon$ from the data, as shown in step $11$ of the Algorithm~\ref{algo:dbscan_core}~\cite{Rahmah_2016}. A $k$-Nearest Neighbor (kNN) algorithm~\cite{Aggarwal_Reddy_2013} is fitted on the combined data of the sensor, $s$, from all the $Q$ tools with $p_{\min}$ as the number of neighbors. Then, the distances of the $k$ nearest neighbors are calculated and sorted in a non-decreasing order. A \emph{knee-point}, i.e., the point on the x-axis, where the distance suddenly change the slope, forming a characteristic knee shape is detected and set as the optimum $\epsilon$. Finally, the clusters are obtained in step $12$ of Algorithm~\ref{algo:dbscan_core}.
\begin{figure}
	\centering
	\includegraphics[width=3in]{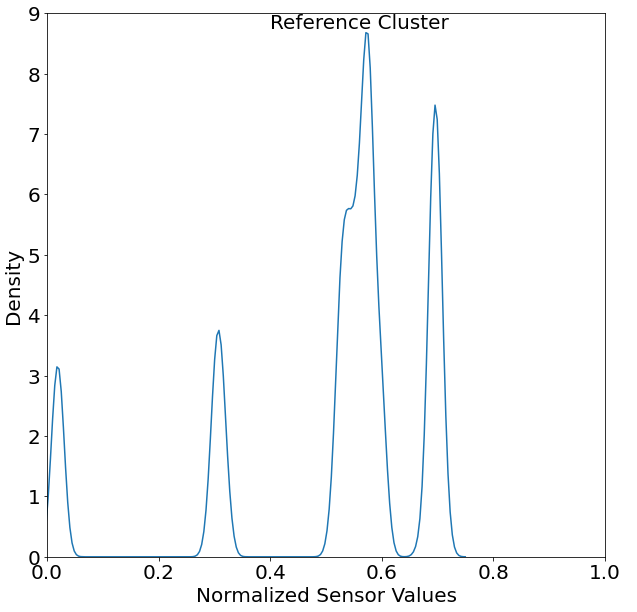}
	\caption{Illustration of the concept of the reference cluster in the density clustering-based algorithm. A kernel density estimator (KDE) is used to obtain a smooth plot as compared to the raw histogram of the min-max normalized sensor values.}
	\label{fig:dbscan_cluster}
	\vspace{-0.2in}
\end{figure}

Post the clustering step, the reference cluster, i.e., the cluster containing the maximum number of points is identified in step $13$. The core idea is that when the tools perform consistently with respect to each other from the perspective of the data of sensor, $s$, \emph{there is only one cluster}, excluding the outliers and other stray noisy data points. In case of one or more tools deviating from an average or the ambient behavior, more than one cluster is formed. An illustration of the concept of the reference cluster for a sensor data is shown in~Fig.~\ref{fig:dbscan_cluster}. The histogram of the normalized sensor values are computed and kernel density estimator~\cite{Hastie_Tibshirani_Friedman_2013} with bandwidth parameter, $B_w = 0.6$, is applied to smooth out the noisy patterns for the illustration purpose. The reference cluster shown in Fig.~\ref{fig:dbscan_cluster} contains the data from $7$ tools and the cluster on the right of the reference cluster contains the data from $3$ tools. Finally, each of the non-reference clusters on the left of the reference cluster contains the data from $1$ tool each.

Finally, we formulate the sensor TTTM difference score as a quantity proportional to the \emph{deviation} in the data of sensor, $s$, of the $q$th tool from the corresponding reference cluster. To this end, we compute the $s$th sensor TTTM difference score denoted by, $\mathbf{d}_{\text{Sensor}}^{\text{DBSCAN}}(s)$, in step $15$ of Algorithm~\ref{algo:dbscan_core} using the centroid, $c_s$, computed in the step $14$. We note that the sensor score is computed such that the worst-case deviation is captured. The sensor level scores are aggregated to obtain the tool level TTTM difference scores as shown in step $17-19$ and assigned to $\mathbf{d}^{\text{DBSCAN}} \in \mathbb{R}^Q$. Finally, the computational complexity of the DBSCAN-based algorithm is $O(Np\log p)$, where $N \triangleq \max_q \{N_q\}$.

\subsection{Statistical Distance Based Scoring}
\label{sec:Statistical_Distance_Based_Scoring}
It is clear from Sec.~\ref{sec:Dataset Preparation} that the T-SUM data per sensor are a collection of statistical quantities. Modeling the underlying phenomenon as an ergodic random process, we view the T-SUM data of $q$th tool for the $s$th sensor as samples drawn from a random variable with an fixed, unknown underlying distribution. Therefore, we view TTTM analysis from the lens of methods which characterize the difference between two (empirical) probability distributions. We use Wasserstein $1$-distance (WD) metric~\cite{Kolouri_Park_2017} denoted by $\text{WD}(\cdot,\cdot)$ to compute the distance between the T-SUM data distribution from a pair of tools.

The proposed computation algorithm is summarized in Algorithm.~\ref{algo:wd_core}.
\begin{algorithm}[t!]
	\caption{Statistical Distance-based Scoring Method}
	\label{algo:wd_core}
	\begin{algorithmic}[1]
		\Require{$\{\mathbf{D}_q\}_{q=1}^{Q}$, $\{\mathcal{S}_q\}_{q=1}^{Q}$}
		\Ensure{$\mathbf{d}^\text{WD} \in \mathbb{R}^Q$}
		\State Compute $\mathcal{S} = \cup_{q=1}^{Q} \mathcal{S}_q$ and set $S \gets |\mathcal{S}|$.
		\State Initialize $\mathbf{d}_{\text{Sensor}}^{\text{WD}} \in \mathbb{R}^{S}$ to an all-zero vector of size $S \times 1$.
		\For {$s \in \mathcal{S}$}
			\State Initialize $\mathbf{d}_s$.
			\For {$q \gets 1~\text{to}~Q$}
				\If {$s \in \mathcal{S}_q$}
					\State $i_s \gets \mathcal{S}_q.{\text{IndexOf}}(s)$.
					\State Update $\mathbf{d}_s \gets \text{Append}(\mathbf{D}_q(i_s, :))$.
				\EndIf
			\EndFor
			\For {$q_1, q_2 \gets 1~\text{to}~Q$}
				\If {$s \in \mathcal{S}_{q_1} \cap \mathcal{S}_{q_2}$}
					\State $d^{\text{WD}} \gets \text{WD}(\mathbf{d}_s(q_1), \mathbf{d}_s(q_2))$
					\State $\mathbf{d}_{\text{Sensor}}^{\text{WD}}(s) \gets \max\{\mathbf{d}_{\text{Sensor}}^{\text{WD}}(s), d^{\text{WD}}\}$
				\EndIf
			\EndFor
		\EndFor
		\State Initialize $\mathbf{D}_{\text{T-T}}^{\text{WD}} \in \mathbb{R}^{Q \times Q}$.
		\For {$q_1, q_2 \gets 1~\text{to}~Q$}
			\State $\mathbf{D}_{\text{T-T}}^{\text{WD}}(q_1, q_2) \gets~\frac{\sum_{s \in \mathcal{S}_{q_1} \cap \mathcal{S}_{q_2}}\mathbf{d}_{\text{Sensor}}^{\text{WD}}(s)}{|\mathcal{S}_{q_1} \cap \mathcal{S}_{q_2}|}$
		\EndFor
		\For {$q \gets 1~\text{to}~Q$}
			\State Set $\mathbf{d}^{\text{WD}}(q) \gets \frac{\sum_{q^\prime = 1}^{Q}\mathbf{D}_{\text{T-T}}^{\text{WD}}(q,q^\prime)}{Q}$
		\EndFor
	\end{algorithmic}
\end{algorithm}
The steps $1-10$ of the Algorithm~\ref{algo:wd_core} collects the required data for each sensor. The step $13$ computes the WD metric between the $q_1$th and $q_2$th tools for each sensor $s \in \mathcal{S}_{q_1} \cap \mathcal{S}_{q_2}$, where $q_1, q_2 \in Q$. In step $14$ of Algorithm~\ref{algo:wd_core}, we set the $s$th sensor TTTM difference score, $\mathbf{d}_{\text{Sensor}}^{\text{WD}}(s)$, as the maximum of all $d^{\text{WD}}$'s across all the pair of tools. Further, a tool-to-tool WD matrix, $\mathbf{D}_{\text{T-T}}^{\text{WD}}$, is computed in step $20$ by aggregating the distances across all sensors and the final $q$th node score, $\mathbf{d}^{\text{WD}}(q)$ is computed in step $23$ by taking the average of $\mathbf{D}_{\text{T-T}}^{\text{WD}}(q,:)$ for each $q \in [Q]$. We note that the computational complexity of the WD-based algorithm is $O\left(N Q^2 \bar{p} \log \bar{p}\right),$ where $\bar{p} \triangleq \sum_q p_q$.

\subsection{Periodogram Based Scoring}
\label{sec:Periodogram_Based_Scoring}
The Periodogram of a time-series (signal) is a non-parametric estimate of the approximate power spectrum of the signal~\cite{Proakis_Dimitris_2006}. The existing methods for comparing time-series data include the auto correlation metric, spectral analysis, model fitting methods, to name a few~\cite{Caiado_Crato_Pena_2009}. The advantage of the Periodogram method is that the long term aging effect induced trend in the data do not affect the difference score characteristics. In contrast, the clustering model could vary unpredictably in the presence of aging effect. Also, WD assumes that empirical probability distribution is fixed over the interval of analysis. Further, least-squares spectral analysis-based estimates obtained by Lomb–Scargle periodogram method do not require uniform sample spacing, as seen in sensor measurements in the Fab with missing data~\cite{Van_Olofsen_1999}. 

The proposed algorithm has similar flow as the Algorithm~\ref{algo:wd_core} with the notations $\mathbf{d}_{\text{Sensor}}^{\text{WD}}$, $d^{\text{WD}}$, $\mathbf{D}_{\text{T-T}}^{\text{WD}}$ replaced by $\mathbf{d}_{\text{Sensor}}^{\text{PD}}$, $d^{\text{PD}}$, $\mathbf{D}_{\text{T-T}}^{\text{PD}}$, respectively and the step $13$ replaced by
\begin{equation}
	d^{\text{PD}} = \left\|\text{Periodogram}\left(\mathbf{d}_s(q_1)\right) - \text{Periodogram}\left(\mathbf{d}_s(q_2)\right)\right\|_2, \nonumber
\end{equation}
where the data is appropriately zero-padded to make the lengths equal and the $\text{Periodogram}(\cdot)$ sub-routine, as the name suggests, computes the periodogram estimate~\cite{Caiado_Crato_Pena_2009, Proakis_Dimitris_2006}. From our simulations, it was found that zero-padding works the best in contrast to the other padding methods~\cite{Proakis_Dimitris_2006}. Lastly, the computational complexity of fast-Fourier transform based periodogram estimator is $O(N Q^2 \tilde{p} \log \tilde{p})$, where $\tilde{p} \triangleq \max_q \{p_q\}$.
 
\section{Univariate Scoring Pipeline}
\label{sec:Univariate Scoring Pipeline}
\begin{figure}[!t]
	\centering
	\includegraphics[width=3in]{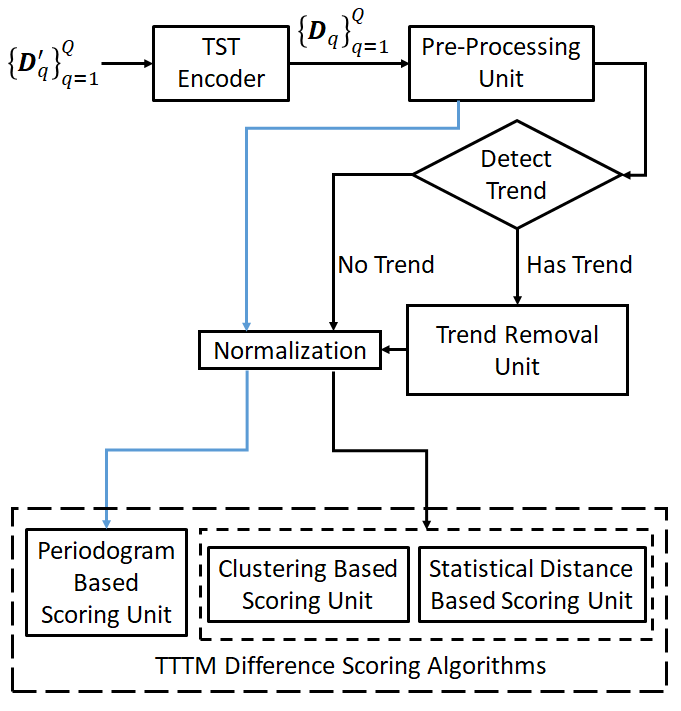}
	\caption{The Flow Diagram of the proposed univariate TTTM Difference Scoring Pipeline. The blue arrows show an independent path for the Periodogram method, where trend detection and removal is not required.}
	\label{fig:tttm_scoring_pipeline}
\end{figure}

In this section, we describe the proposed univariate TTTM difference scoring pipeline with scoring algorithms presented in the previous section, forming its core. The block diagram of the univariate pipeline is shown in Fig.~\ref{fig:tttm_scoring_pipeline}.
\subsection{Preprocessing Unit}
\label{sec:Data_Preprocessing}  
The output of TST-Encoder, i.e., the T-SUM data with $p_q < \tau$ points in the analysis duration are filtered out. Next, the preventive maintenance (PM) schedules in the production line affects the data statistics causing TTTM scores to erroneously spike. For example, part replacement and chamber cleaning operations affect the statistics of the  time-series~\cite{Robert_2007}. Therefore, it is crucial to adjust for this \emph{before/ after (B/A) PM} effect. We analyze the data B/A PM separately to mitigate these side-effects. The PM events for the tool can be obtained from its maintenance logs or can be deduced from the data, by analyzing the behavior of sensors that are \emph{reset} during PM.
\subsection{Trend Detection and Removal Unit}
\label{sec:Trend_Removal}
Next, the part aging induced trend in the data is detected and treated by the \emph{trend removal unit}. We use a non-parametric hypothesis test called Mann-Kendall test~\cite{Scott_Chandler_2011, Gansecki_2010} for detecting monotonic trends in the data.

\subsubsection{Mann-Kendall Test~\cite{Mann_1945, Kendall_1990}}
\label{sec:Mann-Kendall test}
For each sensor, $s \in \mathcal{S}_q$ with index $i_s$ in the $q$th tool, compute
\begin{equation}
	Z_{sq}^\prime = \sum_{\underset{j_1 > j_2}{j_1,j_2=1}}^{p_q}\text{sgn}\left(\mathbf{D}_q(i_s,j_1) - \mathbf{D}_q(i_s,j_2)\right),
\end{equation}
where $\text{sgn}(x)$ denotes the sign function taking values $+1, 0,$ and $-1$ when $x >0$, $=0,$ and $<0$, respectively~\cite{Gilbert_1987}. The quantity $Z_{sq}^\prime$ is then normalized to get the Mann-Kendall test statistic, $Z_{sq}^{\text{MK}}$~\cite{Gilbert_1987}. It is clear that a positive or negative $Z_{sq}^{\text{MK}}$ implies an overall upward or downward trend in the data, respectively. Further, the statistic close to $0$ implies no overall trend. With the significance level $\alpha$ specified, the null hypothesis, $\mathbf{H}_0: \{\text{No~Trend}\}$ is rejected if $Z_{sq}^{\text{MK}} \geq Z_{1-\alpha}$, where $Z_{1-\alpha}$ is the $100(1-\alpha)$th percentile of $\mathcal{N}(0,1)$~\cite{Gilbert_1987}.

Next, we fit three ridge regression models~\cite{Hastie_Tibshirani_Friedman_2013}, each with polynomial features of degree $\in \{1,2,3\}$, respectively. with the regularization constant, $\lambda$. Denote the model with the maximum coefficient of determination~\cite{Hastie_Tibshirani_Friedman_2013} by $f_{sq}^{\text{RR}}$. Remove the trend by $\mathbf{D}_q(i_s,j) \gets \mathbf{D}_q(i_s,j) - f_{sq}^{\text{RR}}(j) + f_{sq}^{\text{RR}}[0]$, where the last term denotes the bias or the constant term in the model.

\subsection{Normalization}
\label{sec:Normalization}
The trend filtered data is then normalized by the min-max normalization in order to bring them to a same scale such that the averaging in step $18$ of Algorithm~\ref{algo:dbscan_core} and the aggregation step in Algorithm~\ref{algo:wd_core} are unweighted. The normalization constants, i.e., the minimum and the maximum values for each sensor $s$ is computed across all the $Q$ tools. This step improves the sensitivity of the TTTM difference score to the variation in the sensor data across the tools. We denote both the original and the normalized, trend subtracted T-SUM data by, $\{\mathbf{D}_q\}_{q=1}^{Q}$, to avoid notation overload.
\subsection{Scoring}
\label{sec:Scoring}
Lastly, the sensor and tool-level TTTM difference scores are computed using (one of) the scoring algorithms discussed in Sec.~\ref{sec:Univariate Scoring Algorithms}. As noted earlier in Sec.~\ref{sec:Periodogram_Based_Scoring}, presence of trend in the data does not affect the Periodogram based scoring algorithm. In this work, we compare the scores obtained by the three proposed methods in terms sensitivity and correlation with the statistical quantities known to affect the consistency of the tool.

\section{Motivation for Multivariate Analysis}
\label{sec:Motivation for Multivariate Analysis}
In summary, the scoring pipeline described in this section computes the sensor scores and then, an aggregation operation is applied to obtain the tool-level scores. In a tool with complex closed control loops, univariate analysis results in one or more issues described as follows.
\begin{enumerate}
	\item Univariate analysis looks at each sensor individually. If several correlated sensors deviate, their individual difference scores add up thereby making the overall score appear larger than the actual deviation. We refer such a deviation as \emph{false difference}.\footnote{We draw parallel to the notion of \emph{false positives} from the area of machine learning and more specifically, classification tasks~\cite{EML_Sammut_Webb_2011, Hastie_Tibshirani_Friedman_2013}} The false differences further results in higher false alarm rates when they are used in tracking the tool consistency and planning (predictive) maintenance activities.
	\item Consider a case, where more than one sensor in a tool deviates, from say, the same in the reference tool, and further, the control loops in the tool adjusts other sensors which might be essentially uncorrelated with the sensor under consideration.\footnote{For instance, if the required set temperature can not be reached in stipulated time due to aging heat actuator, either the pressure in the chamber may be slightly adjusted to a lower value by reducing the gas flow or higher RF power may be supplied to the particles/ electrons to attain the required incident energy.} Then, the univariate analysis-based scores could be potentially higher resulting in \emph{false difference} when in reality, i.e., from the perspective of the yield or the output wafer quality, the tool is behaving consistently with respect to an ambient.
	\item Univariate methods are sensor configuration and type dependent. E.g., if the number of temperature sensors vary between two tools, a way to combine them into a single entity is necessary before the TTTM difference score is computed. Therefore, the univariate methods do not scale well for heterogeneous manufacturing environment. In contrast, as we shall see later, multivariate methods analyze and model the correlation characteristics between the sensors and therefore, overcome this bottleneck.
\end{enumerate}

We have described the issues from the perspective of pair-wise scoring (e,.g., WD and Periodogram-based scores) for succinct illustration of the scenario. However, similar issues occur with the overall tool-level scores since, the averaging operation yields the tool scores from the constituent sensor scores. In a small to mid-sized production lines, where predictive maintenance planning is not a major bottleneck to the line throughput, the above methods can be applied effectively.

In order to overcome the drawbacks discussed above, we propose a multi-variate time-series analysis based scoring pipeline. At the core of the proposed pipeline, lies a graph neural network (GNN)~\cite{Nurul_Yeahia_Ripon_2021, Zhou_Cui_2020}. The GNN-based pipeline learns the graph structure based on the correlation between sensors' behavior, which remain the same for a given recipe. The learned graph structure can be compared between the equipment despite the sensor types varying between the tools and suppliers, up to a large extent. Further, we use the difference or delta change in the scores in the TTTM analysis. The proposed GNN-based TTTM difference scoring pipeline is therefore applicable to heterogeneous settings, which often arises in a modern commercial Fab.

\subsection{Overview of Graph Neural Networks (GNNs)}
\label{sec:Overview of Graph Neural Networks}
As the name suggests, GNNs combine the advantages of both graph representations along with that of the (deep) neural networks. In contrast to the classical grid-based data structures,\footnote{E.g., 1-d array for single time-series and 2-d arrays for multi-variate time series and single-channel images to n-d arrays with $n \geq 3$ for multi-channel images and videos} graphs can represent datasets that inherently consists of relationships among the entities namely social network, citation linkages etc. or datasets like multi-variate time-series acquired from a sensor network with possibly heterogeneous sampling rates or acquisition latency.

In recent times, the concepts from convolutional operator~\cite{Kipf_2016}, attention~\cite{Petar_Yoshua_2017} and message passing~\cite{Gilmer_Riley_2017} have been utilized to either learn to infer from such graph-representation-based datasets or more importantly, to jointly learn the graph representation and perform tasks like anomaly detection~\cite{MTAD_GAT_Zhao_Wang_2020, GTAD_Siwei_Binjie_Zhekang_2022, GDN_Deng_Hooi_2021}. In summary, GNNs have found applications in various fields including bio-sciences, traffic prediction, ad-tech and recommendation systems, social network analysis to name a few~\cite{Dejun_Zhenxing_2021, Wang_2021, Xia_Kai_Liu_2022, Mandal_Maiti_2021}.

\section{Multi-variate Scoring Pipeline}
\label{sec:Multi-variate Scoring Pipeline}
\begin{figure}[!t]
	\centering
	\includegraphics[width=3.4in]{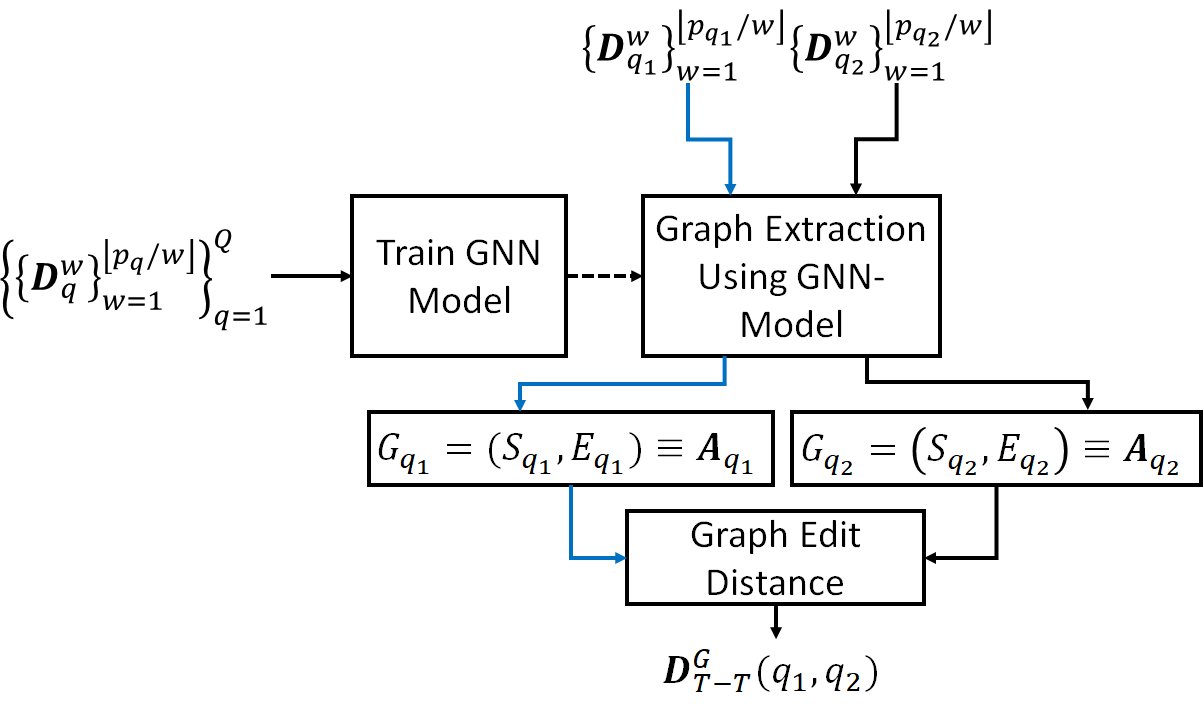}
	\caption{Proposed GNN Pipeline for Pair-Wise TTTM Difference Score Computation. The flow paths marked by the black and blue lines each show that the paths are independent, even though both the data flows through the same blocks.}
	\label{fig:gnn_pipeline}
\end{figure}

The proposed GNN based TTTM Difference scoring pipeline is shown for pair-wise i.e., tool-to-tool score computation in Fig.~\ref{fig:gnn_pipeline}. It is then easy to compute the tool-level scores by using averaging operation similar to that in steps $22-24$ of Algorithm~\ref{algo:wd_core}. As in the univariate pipelines, the T-SUM data from all the sensors in each of the $Q$ tools is collected and pre-processed to remove the intervals where the data is missing\footnote{A probable cause could be node down or recipe/ product change.} in $\{\mathbf{D}_q\}_{q=1}^{Q}$. Recall from Sec.~\ref{sec:Dataset Preparation} that $\mathbf{D}_q \in \mathbb{R}^{N_q \times p_q},$ for each $q \in [Q]$. We note that the column count, $p_q$ can vary across the tools. Without loss of generality, we begin with a formulation where the sensors are same across the tools and therefore, $N_q = N,~\forall~q \in [Q]$. However, it is easy to show that the idea extends to a more general case, as long as the behavior of the tool is captured effectively by the GNN pipeline.\footnote{This situation is analogous to extending, say, a convolutional network on image datasets with different dimensions.}

First, the T-SUM data from the $q$th tool is sliced into multiple windows of equal length, $\omega$, to obtain $\{\mathbf{D}_q^w\}_{w=1}^{\lfloor p_q/\omega \rfloor} \in \mathbb{R}^{N_q \times \omega}$. A \emph{GNN model} is trained with data points, $\{\mathbf{D}_q^w\}$, for all $q \in [Q]$ and $w \in [\lfloor p_q/\omega \rfloor]$ to learn the graph representation of each data point. In the current work, we perform experiments with two GNN architectures: 1) a multi-variate time-series analysis for anomaly detection using graph attention networks (MTAD-GAT)~\cite{MTAD_GAT_Zhao_Wang_2020, GTAD_Siwei_Binjie_Zhekang_2022}, and 2) the graph deviation network (GDN)~\cite{GDN_Deng_Hooi_2021}. We continue to refer this block as the \emph{GNN model} unless a distinction is required to be made.

In order to proceed further, we briefly describe the relevant parts of the MTAD-GAT architecture as follows~\cite{MTAD_GAT_Zhao_Wang_2020}. Each (mini-batch) of the dataset $\{\{\mathbf{D}_q^w\}_{w=1}^{\lfloor p_q/\omega \rfloor}\}_{q=1}^{Q}$ is processed by a $1$-d convolutional layer~\cite{Kiranyaz_Avci_2021} with kernel size, $\kappa$, to extract the local features in each variate yielding the feature matrix, $\mathbf{F}_{wq} \in \mathbb{R}^{N_q \times z^\prime}$, where $z^\prime$ denotes the feature dimension. The extracted features are then processed by two parallel units called time-oriented and feature-oriented graph attention networks (GATs). Since the T-SUM data is a collection of statistics with slowly varying trend, if any, only the feature-oriented GAT is used to obtain the graph representation.

A graph is an ordered pair, $\mathcal{G} = (\mathcal{V}, \mathcal{E})$, comprised of a set of nodes denoted by $\mathcal{V}$ and a set of edges denoted by $\mathcal{E} \subseteq \{\{x,y\} | x,y \in \mathcal{V}, x \neq y\}$~\cite{Biggs_Wilson_1976, Bondy_Murty_2008}. Note that the elements of $\mathcal{E}$ are unordered, denoting that the graph is \emph{undirected}. In our work, the sensors form the nodes and the edges explain the learnt correlation structures between the sensors. Further, a \emph{node feature} denoted by $h_s \in \mathbb{R}^{z}$, $s \in \mathcal{V}$, where $z$ denotes the latent-space dimension is learnt from the feature matrix, $\mathbf{F}_{wq}$ using GATs and imposed on the graph. The attention weights in GAT~\cite{MTAD_GAT_Zhao_Wang_2020} determine the \emph{arrangement} of the node features in an appropriate metric space, and hence determines the graph structure through the cosine distance distributions. An edge $e = \{s_1, s_2\} \in \mathcal{E}$ is defined between $s_1, s_2 \in \mathcal{V}$ if the cosine distance~\cite{Hastie_Tibshirani_Friedman_2013} between the node features, $h_{s_1}, h_{s_2}$ are above a certain threshold, $\tau_g$~\cite{MTAD_GAT_Zhao_Wang_2020}. An undirected graph $\mathcal{G}$ can be equivalently represented by an adjacency matrix, $\mathbf{A} \in \{0,1\}^{|\mathcal{V}| \times |\mathcal{V}|}$ such that $\mathbf{A}(i_{s_1},i_{s_2}) = 1$ if $\{s_1, s_2\} \in \mathcal{E},$ where $i_{s_1}$ and $i_{s_2}$ are the indices of $s_1$ and $s_2$ in $\mathcal{S}_q$. The nature of the adjacency matrix dictates that the edges are \emph{unweighted}~\cite{Gilbert_Kepler_2011}.

The node features from each window are then used to learn the time dependencies using a gated recurrent unit (GRU) network with $L_{\text{GRU}}$ hidden layers and $H_{\text{GRU}}$ units. Lastly, the task of learning the graph representation for a multi-variate time-series is driven by two tasks namely, time-series forecasting and data reconstruction~\cite{MTAD_GAT_Zhao_Wang_2020}. The sum of forecasting and reconstruction loss is back-propagated to learn the weights of the entire GNN Model~\cite{MTAD_GAT_Zhao_Wang_2020}.

We note that in the presence of yield or other useful metrology information, a classification or regression task can drive the weight updates as opposed to forecasting or reconstruction losses. At the same time, if such metrology data is available, it can be argued from the discussion in~\ref{sec:Introduction} that the tool consistency check can be performed using the said metrology data. However in practice, metrology is not performed on all the materials passing through the node. Further, the final yield is available only once all the processing steps are completed thereby introducing a latency of few weeks to couple of months. In addition, final yield comprises of all the variations across the entire spectrum of tools the material has been processed with and hence, not suitable for tool-focused scoring.

Post training, the learnt GNN model is used to obtain a graph representation between any $q_1$th and $q_2$th tools. A threshold denoted by $\tau_g$ is used on the cosine scores between the node features, below which the edges are trimmed to avoid noise in the data contributing to the edges. We treat $\tau_g$ as a hyper-parameter in our experiments. We vary $\tau_g$ in range $[0.5, 1.0]$ and compare the characteristics of the TTTM scores (see Sec.~\ref{sec:Multivariate Results}).

The adjacency matrices (of the graph) obtained for each $w \in [\lfloor p_q/\omega \rfloor]$ given a $q \in [Q]$ is averaged across all the windows to obtain an average adjacency matrix denoted by $\mathbf{A}_q$, for the $q$th tool. The (normalized) graph edit distance is computed between the $q_1$th and $q_2$th tool-pair as
\begin{equation}
	\mathbf{D}_{\text{T-T}}^{\mathcal{G}}(q_1, q_2) = \frac{\sum_{i,j=1}^{N}|\mathbf{A}_{q_1}(i,j) - \mathbf{A}_{q_2}(i,j)|}{E_{\max}},
\end{equation}
where $E_{\max}$ denotes the maximum number of edges possible, learnt from the training data. We can extend this metric to a case where the sensors are different by defining the summation over $S_u = |\mathcal{S}_{q_1} \cup \mathcal{S}_{q_2}|$.\footnote{The formulation used is analogous to that of any natural language processing (NLP) or text mining application, where the goal is to transform a text into its vector representation like CountVectorizer, TF-IDF, Word2Vec, to name a few~\cite{NLP_Brownlee_2017}.} Finally, the tool-level TTTM score can be computed using an averaging operation similar to that used in step $23$ of Algorithm~\ref{algo:wd_core}.

\begin{figure}
	\centering
	\includegraphics[width=3.5in]{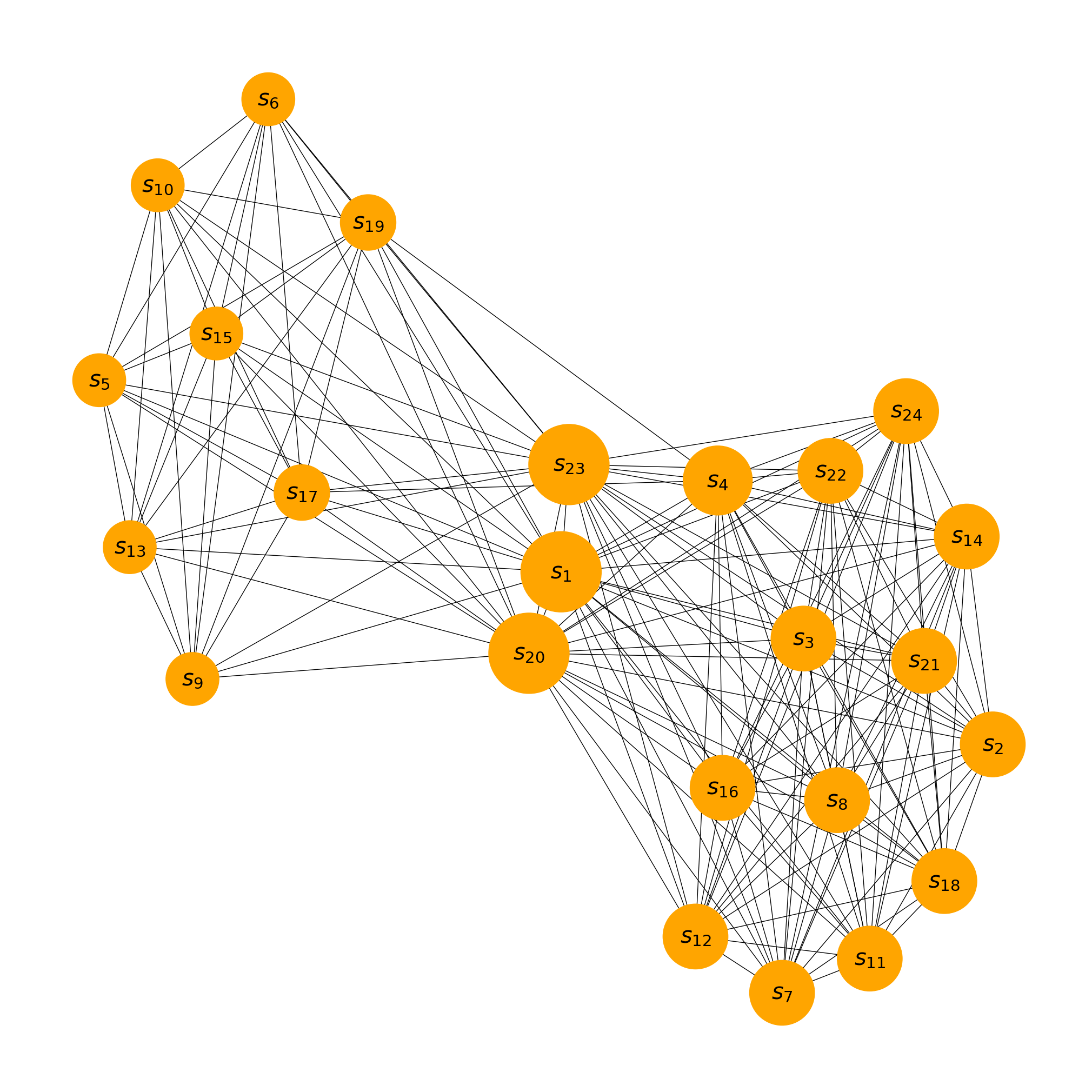}
	\caption{A sample graph representation learnt from the tool's T-SUM data using the proposed GNN-based multi-variate TTTM difference scoring pipeline.}
	\label{fig:mtad_gat_graph_sample}
\end{figure}

We conclude this section by illustrating a sample graph representation of a tool learnt using the proposed pipeline in Fig.~\ref{fig:mtad_gat_graph_sample}. The sensors (nodes in the graph) with higher number of degree are enlarged to provide a better representation of the inter-sensor relationships. For visualization clarity, the edge weights and node features are not shown in the figure. 

The MTAD-GAT architecture can be replaced by an architecturally simpler GDN, which has lower number of parameters to train. As we shall see in the simulation section, the GDN performance is slightly lower but due to its low architectural complexity is attractive when the dataset size is small. Finally, the computational complexity of the GNN-based multi-variate scoring pipeline is $O\left( Q(\omega N^2 Z + \omega N^2) + Q^2 N^2 \right)$.

\section{Results and Observations}
\label{sec:Results_and_Observations}
\subsection{Univariate Algorithms}
\label{sec:Univariate Results}
\begin{figure}[!t]
	\centering
	\includegraphics[width=3.4in]{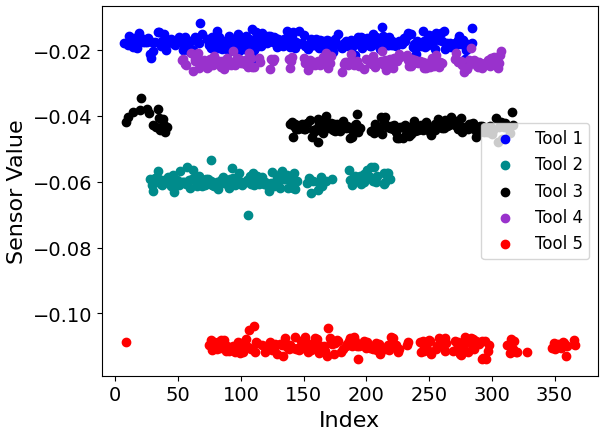}
	\caption{A sample sensor data with high TTTM Difference Score = $0.822$.}
	\label{fig:high_tttm}
\end{figure}

\begin{figure}[!t]
	\centering
	\includegraphics[width=3.4in]{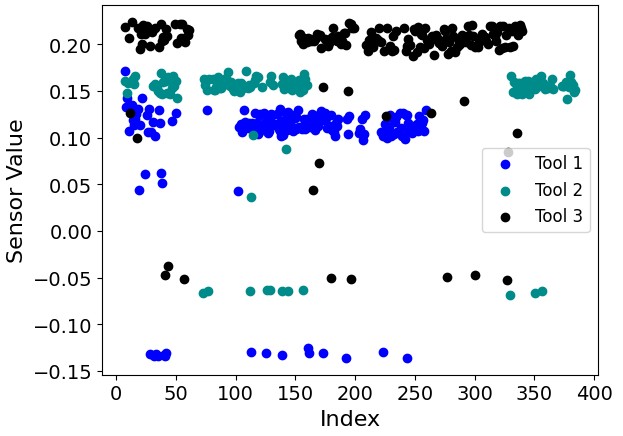}
	\caption{A sample sensor data with low TTTM Difference Score = $0.142$.}
	\label{fig:low_tttm}
\end{figure}

In this section, we present the results obtained by the application of the proposed univariate and multivariate pipelines to our experimental dataset.\footnote{The data is sourced from one of the Samsung Fab and is proprietary. Appropriate masking logic has been applied to the sensor names, the tool names. It is ensured that the relevant characteristics of the data remains same in order to retain the validity of the results and observations.} The results are generated using the following parameters in our pipeline. The dataset contains the sensor data from $Q = 9$ tools and $N_q = 24,~\forall~q \in Q$ sensors, corresponding to $2,419$ wafers process runs in the period of $1$ month. The minimum number of points required by the DBSCAN-based method, $p_{\min} =2$. The threshold on the minimum number of points required for the analysis, $\tau = 10$ is used. Further, the significance level in the Mann-Kendall test, $\alpha = 0.05$, along with the regularization constant in the ridge regression, $\lambda = 1$, is used. Two sample sensor T-SUM plots along with the corresponding univariate pipeline computed TTTM difference scores, using the clustering-based pipeline, are illustrated in Fig.~\ref{fig:high_tttm} and Fig.~\ref{fig:low_tttm}.

From the Fig.~\ref{fig:high_tttm}, we note that the behavior of the sensor in the tool whose data is marked in the red-color is significantly different from that of the other tools, i.e., blue, orchid, and cyan-colored plots. At the same time, the variation across the tools are significant as compared to the variance per tool. Therefore, we obtain a high sensor TTTM difference score of $0.822$. From Fig.~\ref{fig:low_tttm}, we note that the sensor behavior is relatively similar taking the per-tool variance into consideration, excluding the stray or low frequency outlier around in the range $[-0.15, 0.05]$ yielding a low score of $0.142$.
\begin{table}[t]
	\renewcommand{\arraystretch}{1.3}
	\centering
	\caption{Correlation Analysis}
	\label{tab:corr_scores}
	\begin{tabular}{|c|c|c|c|}
			\hline
			\textbf{Statistics} & \textbf{DBSCAN} & \textbf{WD} & \textbf{PD}\\			
			\hline
			\textbf{Variance} & 0.969 & 0.947 & 0.712	\\
			\hline
			\textbf{Number of Modes} & 0.525 & 0.547 & 0.266 \\
			\hline
	\end{tabular}
\end{table}

Next, we analyze the correlation of the sensor TTTM difference scores with the statistical quantities, which are known to affect the process consistency namely, the variance of the data across the tools and the number of modes in the data distribution. The number of modes can be estimated by using level shift method on the cumulative distribution function of the sensor T-SUM data~\cite{Tsay_1988}.

Consider computing the variance of the sensor T-SUM across all the nodes. If the variance is low, it is likely that the TTTM difference score is also low. We note that identification of such features, i.e., sensor parameters, can yield a nice rule-based approach to compute TTTM score. Unfortunately, such rule-based model construction is not practical as exhaustive set of parameters that affects the score is unknown along with the contributing factors/ weights. 

The Spearman correlation coefficients~\cite{Pearson_Spearman_2008} are tabulated in Table~\ref{tab:corr_scores}. We use Spearman rank correlation since the scores are not distributed according to a Gaussian distribution. From the table, we can observe that the correlations are positive showing a direct relation between the statistical quantities and the scores computed from our pipeline. We observe from the Table~\ref{tab:corr_scores} that both the clustering-based and statistical distance algorithms show high correlation coefficient of $> 0.95$ and $> 0.5$ with respect to the variance and the number of modes, respectively. In contrast, the Periodogram method had lower correlation of $\approx 0.71$ and $\approx 0.266$ with respect to the variance and the number of modes, respectively.

We obtained domain knowledge based feedback that, indeed the clustering method-based pipeline is better, although an additional trend processing unit is necessary. Therefore, we suggest applying the clustering method for univariate TTTM difference score computation and analysis followed closely by the WD method.

\subsection{Multivariate Algorithms}
\label{sec:Multivariate Results}
In this subsection, we present the results obtained by our multivariate pipeline on the same dataset described in Sec.~\ref{sec:Univariate Results}. We start by specifying the parameters used by the pipeline. We set the window size, kernel size, dropout and learning rate of gradient descent algorithm as $\omega=7$, $\kappa=3$, $p_{do} = 0.2$, and $\eta=10^{-3}$, respectively. Further, $L_{\text{GRU}}=1$ and $H_{\text{GRU}} = 50$ are used. A leaky-ReLU activation function with the negative slope $= 0.2$ is used in the GAT along with an embedding space dimension, $z=50$. In contrast, the GDN-based model requires a higher embedding space dimension, $z=128$. The forecast network and the reconstruction network each has $1$ layer of size $50$. Finally, the GNN Model is trained with $1000$ epochs with mini-batch size of $16$.
\begin{table}[t]
	\renewcommand{\arraystretch}{1.3}
	\centering
	\caption{Correlation Between the GNN-based Pipeline and Univariate (DBSCAN and WD) Scoring Pipelines.}
	\label{tab:corr_mtad_gat}
	\begin{tabular}{|c|c|c|c|c|}
		\hline
		\multirow{2}{*}{\textbf{$\boldsymbol{\tau}_\textbf{g}$}} & \multicolumn{2}{c|}{\textbf{MTAD-GAT-Based}} & \multicolumn{2}{c|}{\textbf{GDN-Based}} \\
		\cline{2-5} & \textbf{DBSCAN} & \textbf{WD} & \textbf{DBSCAN} & \textbf{WD}\\			
		\hline
		$0.5$ & 0.73 & 0.39 & 0.4 & 0.62\\
		\hline
		$0.8$ & 0.76 & 0.485 & 0.37 & 0.54\\
		\hline
		$0.9$ & 0.754 & 0.488 & 0.63 & 0.61\\
		\hline
		$0.95$ & 0.69 & 0.488 & 0.75 & 0.56\\
		\hline
	\end{tabular}
\end{table}

We tabulate the correlation results in Table~\ref{tab:corr_mtad_gat} across various thresholds, $\tau_g \in \{0.5, 0.8, 0.9, 0.95\}$. From table~\ref{tab:corr_mtad_gat}, we observe that the correlation between the pair-wise scores and that obtained by WD algorithm based pipeline shows peak at $\tau_g = 0.9$ and $0.95$ whereas, DBSCAN shows a peak when $\tau_g \geq 0.9$. In summary, we find that $\tau_g = 0.9$ is a good threshold for our dataset. Further, we note that irrespective of the values of $\tau_g$ considered in the range $[0.5, 1.0)$, the GNN Model-based pipeline scores show a positive correlation with that obtained by univariate methods like DBSCAN and WD-based pipelines. Recall that the DBSCAN-based pipeline computes the scores at the tool level whereas, WD-based pipeline computes an intermediate pair-wise i.e., tool-to-tool scores. Therefore, comparing the multivariate pipeline scores with both DBSCAN and WD-based pipeline scores serves as sanity check at both pair-wise and tool-levels. As a closing remark to the comparison exercise, we suggest applying a weighted average of the scores from both MTAD-GAT and GDN pipelines, where the weights are fine-tuned on the field to reduce the false alarm and mis-detection.

Finally, we characterize the statistical behavior of the pair-wise scores as $\tau_g$ is varied for the MTAD-GAT-based pipeline averaged over $100$ Monte-Carlo runs in Table~\ref{tab:gnn_scores}.
\begin{table}[t]
	\renewcommand{\arraystretch}{1.3}
	\centering
	\caption{Maximum, Minimum and Standard Deviation of the Unnormalized TTTM Difference Scores Obtained from the MTAD-GAT-based pipeline across various $\tau_g$ averaged over $100$ Monte-Carlo Runs.}
	\label{tab:gnn_scores}
	\begin{tabular}{|c|c|c|c|}
		\hline
		\textbf{$\boldsymbol{\tau}_\textbf{g}$} & \textbf{Maximum Scores} & \textbf{Minimum Scores} & \textbf{Standard Deviation}\\			
		\hline
		$0.5$ & 190.08 & 36.58 & 35.39\\
		\hline
		$0.8$ & 170.14 & 37.97 & 31.51\\
		\hline
		$0.9$ & 160.01 & 34.19 & 29.82\\
		\hline
		$0.95$ & 150.62 & 32.97 & 27.79\\
		\hline
		$0.98$ & 135.58 & 30.60 & 24.89\\
		\hline
		$1.0$ & 101.05 & 1.09 & 22.28\\
		\hline
	\end{tabular}
\end{table}
From Table~\ref{tab:gnn_scores}, we observe that the maximum and minimum unnormalized pair-wise TTTM scores decrease as $\tau_g$ increases. Further, a sharp decrease in the minimum unnormalized score is noticed when $\tau_g = 1.0$ is used. Higher thresholds reduce the number of edges in the graph representation thereby making the adjacency matrix relatively more sparse. As a result, there is a lower chance of seeing differences in the entries. This aspect can be attributed to the decrease in the average scores. Also, the standard deviation between the scores reduces as $\tau_g$ increases thereby illustrating (an indirect) inverse relationship between the sensitivity i.e., how significant is a unit increase or decrease of the scores to the threshold. Based on the business needs including the bound on the false alarm rates for planning maintenance slots, the threshold can be set on the field easily. 

\subsection{Standard Operating Procedure: Tool Consistency Tracking}
\label{sec:SOP Tool Consistency Monitoring}
\begin{figure}[!t]
	\centering
	\includegraphics[width=2.9in]{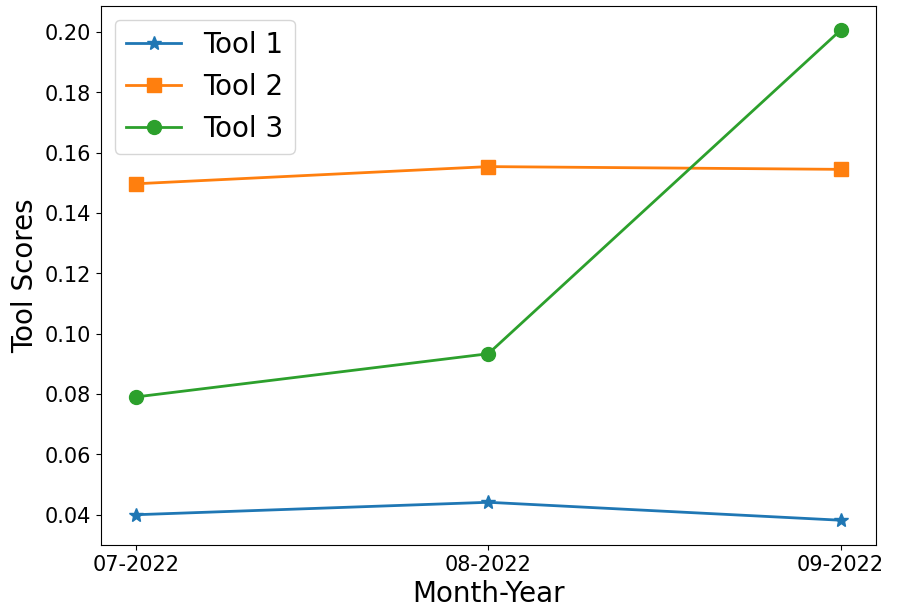}
	\caption{The trend of the tool-level TTTM difference scores for $3$ Etch tools observed over a duration of $3$ months.}
	\label{fig:tool_trend}
\end{figure}

In this section, we illustrate a standard operating procedure for consistency monitoring at tool-level and drill-down steps with an example. To this end, we acquire the T-SUM data from $3$ Etch tools, each containing $20+$ sensors of various types~\cite{Han_Min_Ma_2023, Kevin_Turner_Jackson_2000}: \emph{Chiller Flow}, \emph{Helium Gas Flow}, \emph{Clamp Pressure}, \emph{Temperature}, \emph{Valve position of the mass flow controller (MFC)}, to name a few over $3$ months duration. The trend of the tool scores over each month is visualized in a dashboard. A sample plot from the said dashboard is shown in Fig.~\ref{fig:tool_trend}. 

\begin{figure}[!t]
	\centering
	\includegraphics[width=2.9in]{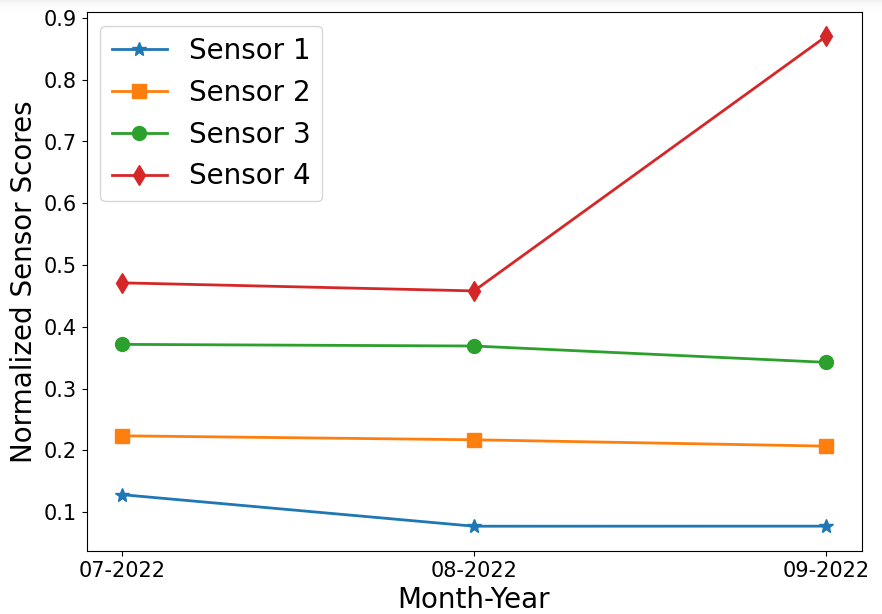}
	\caption{The trend of the normalized sensor-level TTTM difference scores for $4$ sensors observed over a duration of $3$ months. The plots of the other sensors which do not contain the information required for our purpose in this section, are not shown for visualization clarity. All the plots shown belong to a certain \emph{Helium gas flow} sensors.}
	\label{fig:sensor_trend}
\end{figure}

From the Fig.~\ref{fig:tool_trend}, we observe that \emph{Tool 3} shows a relatively higher deviation in the month of Sept., 2022. The normalized sensor scores for the Tool $3$, illustrated in Fig.~\ref{fig:sensor_trend} was observed. It was found out that the \emph{Sensor 4}, a specific type of sensor which measures the Helium Gas flow in the chamber show a deviation. In a typical Fab, the relevant information can be monitored over a dashboard with statistical process chart (SPC) out-of-spec limit rules applied on the tool-scores and sensor-score to raise an alarm in case of a deviation~\cite{Sarfaty_Paik_Parikh_2002}.

\section{Conclusion}
\label{sec:Conclusion}
We considered the problem of quantification of process or the functional consistency of an equipment, the equipment chamber and the sub-modules of the equipment using tool-level and sensor-level tool-to-tool matching (TTTM) analysis, respectively. Specifically, we proposed novel univariate TTTM difference scoring pipelines based on three core algorithms namely: density-based clustering, statistical distance and periodogram. The sensitivity and correlation-based analysis of the scores obtained from the univariate methods with statistical quantities known to affect the consistency of the process were conducted and the results presented showed the effectiveness of our pipelines. In addition, we proposed a novel graph neural network (GNN) based pipeline for the multivariate analysis based TTTM difference scoring pipeline. The correlation analysis with the univariate methods showed that the multivariate results show the same characteristics and therefore are effective in explaining the consistency state of the tool. At the same time, the multivariate pipelines can overcome the issues of false difference, which typically affect the univariate pipelines. As part of the future work, one could use machine-learning techniques like auto-encoders and other representation learning architectures to convert raw time-series to T-SUM data~\cite{Kortmann_Moritz_Mark_2021}. In addition, extension of the GNN-based multivariate pipelines to model the state of the tools across multiple process recipes and product mix, which are common scenario in commercial Fabs is useful. 

\nocite{*}
\bibliographystyle{IEEEtran}
\bibliography{IEEEabrv,References_Y2022_P2.bib}

\vfill

\end{document}